\documentclass[conference]{IEEEtran}
\IEEEoverridecommandlockouts
\usepackage{cite}
\usepackage{amsmath,amssymb,amsfonts}
\usepackage{algorithmic}
\usepackage{booktabs}
\usepackage{array}
\usepackage{graphicx}
\usepackage{textcomp}
\usepackage{xcolor}
\usepackage{float}
\usepackage{flafter}
\usepackage{multirow}
\def\BibTeX{{\rm B\kern-.05em{\sc i\kern-.025em b}\kern-.08em
    T\kern-.1667em\lower.7ex\hbox{E}\kern-.125emX}}

\begin{document}

\title{From Video to Vision: Learning Generalizable Humanoid  Carrying from Human Motion\\
}

\author{
\IEEEauthorblockN{
Jianming Lu\textsuperscript{1,2},
Ziluo Ding\textsuperscript{2\dag}\thanks{\textsuperscript{\dag} Team Leader.},
Zhen Zhao\textsuperscript{2},
Shuyang Jin\textsuperscript{2},
Han Wen\textsuperscript{2},
Xinrun Xu\textsuperscript{2},
Zhiyuan Xu\textsuperscript{2}
}
\IEEEauthorblockA{
\textsuperscript{1}Institute of Automation, Chinese Academy of Sciences\\
\textsuperscript{2}X-Humanoid\\
Email: lujianming2025@ia.ac.cn, elf.ding@x-humanoid.com
}
}

\maketitle

\begin{abstract}
In the field of whole-body interaction between humanoid and scene objects, 
achieving generalizable and flexible object interaction across diverse scenarios has remained a long-standing challenge. 
In object carrying tasks, the humanoid can only lift objects from designated locations along the reference trajectory, 
resulting in poor spatial generalization. Moreover, existing policies can only be evaluated inside a motion-capture studio. 
To address these limitations, we propose \textbf{MOVER}, 
a multi-stage, Teacher-Assistant-Student, RL training framework for humanoid object carrying. 
We first collect data from videos and employ GENMO and GMR to retarget the video data, 
combining this with data augmentation techniques to generalize the data across objects of varying shapes and sizes. 
In the \textbf{first stage}, building upon the DeepMimic \cite{peng2018deepmimic} motion-training framework, 
we train the humanoid to transport objects from specified locations along designated trajectories. 
In the \textbf{second stage}, we relax the tracking of the reference motion and adjust the reward so that the humanoid learns to transport objects from different positions. 
In the \textbf{third stage}, we apply a DAgger-PPO distillation method to distill away the proprioceptive position observations and reference joint position of the humanoid. 
This enables the robot to directly identify the object's position and orientation through the RGB-D camera mounted on its head, thereby completing the carrying task.
We design a series of simulation experiments to compare the success rates across stages. 
We further transfer the policy to a real humanoid, successfully transporting objects at varying positions in a motion-capture-free setting.
\end{abstract}
    

\section{Introduction}


Enabling humanoid to interact with surrounding terrain and scene objects during whole-body motion control has attracted increasing attention from the research community.
Meanwhile, the demand for humanoid in warehousing and logistics scenarios is growing rapidly.

Using humanoid carrying as a representative application. For objects with varying placements and states, existing methods generally rely on training with large amounts of reference trajectory data. 
At inference time, prior reference motion trajectories are required as input to the policy network to execute interaction motions for objects in different states \cite{wuSUGARScalableHumanVideoDriven2026, yang2025omniretarget}. 
However, such training data can hardly cover every possible humanoid-object interaction posture.
When the agent encounters trajectories that substantially differ from those seen during training, real-world deployment tends to produce unstable humanoid motions.
Moreover, most existing methods can only be evaluated within mocap studios, 
where optical mocap is used to obtain the poses of both the humanoid and the object, 
severely restricting their applicability to production environments.

To address these issues, we propose \textbf{MOVER},
a multi-stage humanoid learning framework for carrying task.
We first extract reference motion data from videos of humans carrying objects, 
retarget the motion data to the humanoid's trajectories, 
and employ data augmentation strategies to generate a large volume of training trajectories.
These reference motions are integrated into our RL training framework.
In the \textbf{first stage}, we train a 29Dof Tiangong 3.0 humanoid to strictly track reference trajectories while carrying objects, yielding a teacher policy.
In the \textbf{second stage}, our objective is to transition the humanoid from "imitating fixed motions" to "autonomously completing carrying tasks."
To this end, we relax the strict trajectory tracking requirements. Instead, we strengthen the rewards for physical interaction with objects and for tracking feedback during the carrying process. This enables the humanoid to learn to locate and transport objects placed at different locations.
In the \textbf{third stage}, we apply the \textit{DAgger-PPO} \cite{zhang2025distillation} distillation method, using the Stage-2 policy as the assistant policy, 
to distill away the humanoid's proprioceptive position observations and reference joint position information.
At deployment, the humanoid perceives the relative position and pose of the object directly through its onboard vision system, enabling vision-based object tracking and carrying.

However, the resulting student policy still retains reference information about the humanoid's anchor pose.
If this information is further distilled away, the humanoid cannot determine the direction of the destination after picking up the object.
To completely eliminate reference motion observations, we design a \textit{LSTM} neural network trained with supervised learning on large-scale motion data, 
which takes as input the object's pose and the target position, and outputs the reference orientation of the humanoid.
The final deployed policy's observations contain only the humanoid's proprioceptive information and the object's relative pose information.
Experimental results demonstrate that, on the object-carrying task, our method achieves a success rate of 93\% in simulation, surpassing existing approaches.
The distilled student policy transfers successfully to a physical humanoid, where it can complete carrying tasks in general, 
mocap-free scenes and exhibits a broader object-position generalization range than most existing humanoid object-carrying methods.

The main contributions of this paper are summarized as follows:
\begin{enumerate}
  \item We propose a multi-stage humanoid learning framework that, given a single reference motion, effectively enables the humanoid to learn to grasp and transport objects at varying positions.
  \item We design an neural network trained with supervised learning on reference motion data to learn the mapping from object-humanoid relative pose to humanoid reference rotations, completely eliminating reference motion observations from the deployed policy's observations.
  \item We designed multiple sim-to-real experiments and successfully verified that our policy can complete object-carrying tasks for objects at different positions using vision alone.
\end{enumerate}


\section{Related Work}
\subsection{Human Motion Recovery and Retargeting}
Human motion is an effective source of reference data for humanoid control, but the embodiment gap prevents direct transfer. 
Whole-body geometric retargeting matches selected body poses through inverse kinematics \cite{darvish2019whole,
he2024learning,luo2023perpetual,ProtoMotions, araujo2025gmr}.  
He et al. \cite{he2024learning} and Luo et al. \cite{luo2023perpetual} instead fit human-body representations to the humanoid, 
while Tessler et al. \cite{ProtoMotions} uses scaled Cartesian targets. 
These methods can introduce foot sliding, penetration, or self-collision 
when body proportions and contacts are not handled well. GMR reduces these artifacts through rest-pose alignment, 
non-uniform local scaling, and two-stage constrained optimization \cite{araujo2025gmr}.

Video-based pipelines reduce the need for motion-capture systems. 
SFV learns physical skills from monocular video \cite{2018-TOG-SFV}, and Peng et al. \cite{shen2024world} recovers world-grounded human motion under camera movement. 
Recent systems extend this idea to humanoid-scene and humanoid-object interaction\cite{allshire2025visualimitationenablescontextual,
weng2025hdmi, yang2025omniretarget, wang_humanx_2026}. 
Li et al. proposed GENMO \cite{genmo2025} to unify motion estimation and generation, 
which can provide robust priors under occlusions. We use GENMO for video-only human motion recovery and GMR\cite{araujo2025gmr} 
for humanoid retargeting, then augment the object and interaction states for broader spatial and geometric coverage.
\begin{figure*}[htbp]
    \centerline{\includegraphics[width=17cm]{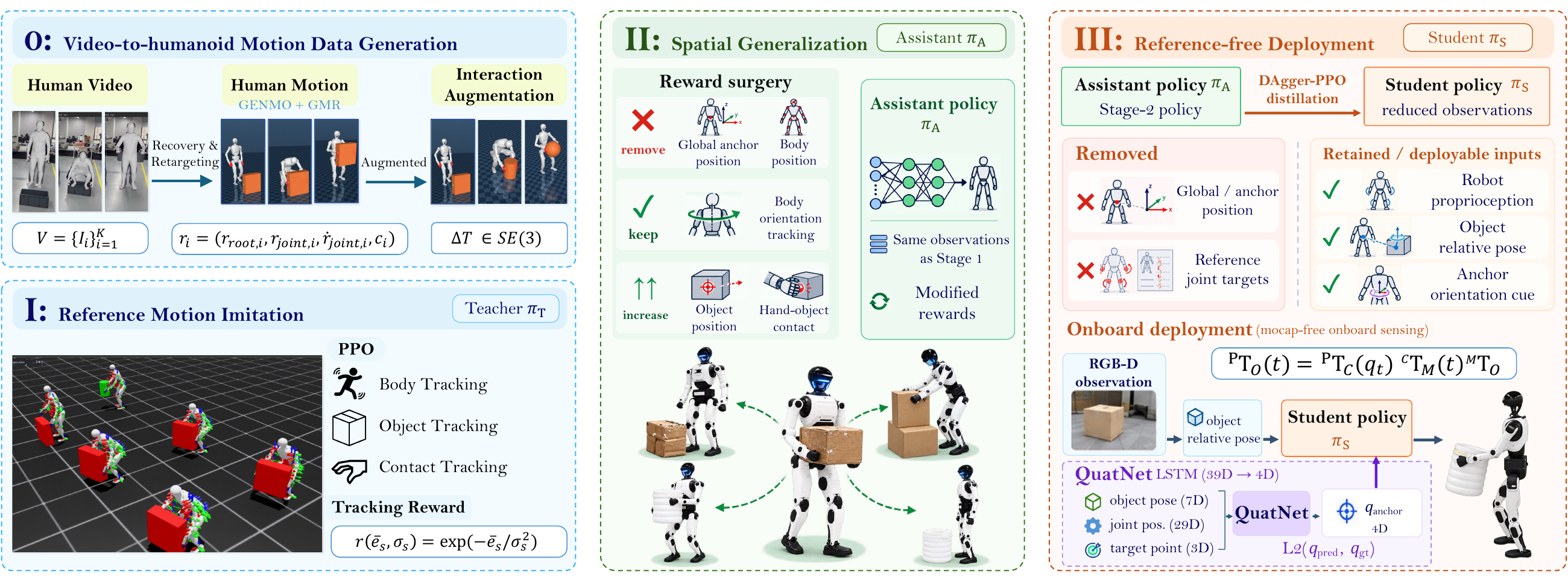}}
    \caption{MOVER Overview.(0) Generating augmented humanoid motion data from human video data. (1) Imitation learning training for humanoid object carrying. (2) Two-stage training to improve carrying generalization. (3) In the third stage, DAgger-PPO distillation is used to enable the humanoid to complete object carrying relying purely on vision.}
    \label{fig:overview}
\end{figure*}

\subsection{Humanoid Whole-Body Interaction}
Humanoid whole-body motion skill learning has achieved significant progress in recent years.
Some works use deep reinforcement learning to learn object interaction, they need to design complex reward functions in multiple stages, 
and due to the lack of reference actions, the actions learned by the humanoid are significantly different from normal human actions \cite{zhang2505falcon,zhang2024wococo}.
Methods such as ASAP \cite{he2025asap} and BeyondMimic \cite{liao2025beyondmimic} provide standardized imitation learning frameworks for humanoid, 
enabling high-fidelity reproduction of retargeted human motions on physical platforms.
Nevertheless, these approaches have not been extended to humanoid-object interaction scenarios.

HDMI pioneered whole-body interactive imitation learning, 
successfully demonstrating tasks such as door opening, box carrying, and chair unfolding \cite{weng2025hdmi}.
The recent OmniRetarget introduces a retargeting pipeline for scene-interactive motions, 
producing high-quality reference motions that achieve excellent imitation with minimal reward engineering \cite{yang2025omniretarget}.
Both methods, however, heavily rely on reference motions and struggle to generalize to objects in varied configurations.
Furthermore, OmniRetarget does not incorporate object position information in the actor observation, 
limiting objects to fixed poses matching the reference motion during testing.
Our approach employs a multi-stage RL training framework that substantially reduces dependence on reference motions, 
enhancing generalization for object carrying and enabling effective handling of objects in diverse scenes and poses during real-world deployment.
\section{Method}
\subsection{Video-to-humanoid Motion Data Generation}
Direct video retargeting produces only one humanoid–object trajectory, limiting the variation in object geometry and placement. We augment the object configuration while preserving the demonstrated hand–object relationship, and adapt the upper-body motion using constrained inverse kinematics. This produces contact-consistent references across different object sizes, poses, and support heights using only a small set of videos.

Given an RGB video $V = \{I_i\}_{i=1}^K$ of a person carrying a object, we use the estimation mode of GENMO \cite{genmo2025} to recover the human motion.
Only video-aligned conditions are enabled; text, music, and keyframe inputs are masked out.
The estimated state is $h_i = (h_{\text{root},i}, h_{\text{joint},i})$, where $h_{\text{root},i}$ is the world-grounded root pose 
and $h_{\text{joint},i}$ contains the SMPL joint rotations \cite{SMPL:2015}.
Sequence-level estimation separates camera and human motion and remains stable under partial object occlusion.

We map $\{h_i\}$ to the humanoid with GMR \cite{araujo2025gmr}. Human and humanoid key bodies are aligned in their rest poses,
and non-uniform local scaling accounts for different body proportions.
Two-stage inverse kinematics then tracks link orientations and key-body positions under the humanoid joint limits.
Each frame is initialized from the previous solution to keep the motion smooth.
This gives a humanoid reference $r_i = (r_{\text{root},i}, r_{\text{joint},i}, \dot{r}_{\text{joint},i}, c_i)$, including the floating-base pose, joint state, and contacts.

We further augment the object geometry, size, and initial pose.
For each reference, we sample a object mesh, anisotropic scale, and a valid pose perturbation $\Delta \mathbf{T} \in \mathrm{SE}(3)$.
A hand anchor is defined from the two palm poses.
During grasping and carrying, the object pose is propagated by $\mathbf{T}_i^o = \mathbf{T}_i^a (\mathbf{T}_c^a)^{-1} \mathbf{T}_c^o$, 
which preserves the object-hand transform at the contact frame $c$. 
For each augmented object, constrained inverse kinematics adjusts the arms and torso while enforcing palm contact, 
joint limits, collision clearance, and support constraints. 
The approach segment is shifted and smoothly connected to the refined contact pose. 
The resulting references cover different object shapes, sizes, positions, orientations, and support heights. 

\begin{figure*}[htbp]
    \centerline{\includegraphics{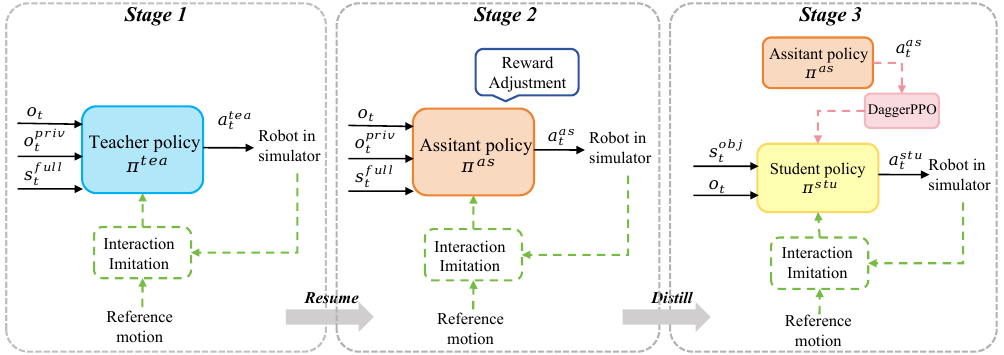}}
    \caption{The training pipeline of MOVER.
    In the first stage, the teacher model is trained using the full observation set.
    In the second stage, the reward function is adjusted and the teacher policy is resumed to obtain the assistant policy.
    In the third stage, distillation is performed using the DAgger-PPO algorithm, yielding the student policy.
    The student takes the object's relative observations and the humanoid's proprioception as input, 
    and can be directly deployed on the real humanoid.
    }
    \label{fig:2}
\end{figure*}

\subsection{RL Training Setup}
\subsubsection{Stage 1: Teacher Policy Training for Reference Motion Tracking}
After processing video data to obtain enhanced motion trajectory data, 
this RL training is modeled as a Partially Observable Markov Decision Process (POMDP).
The core training objective is for the humanoid to learn to transport objects to designated positions in the environment.
We employ PPO \cite{schulman2017proximal} as the reinforcement learning algorithm.
The MDP observation space encompasses the humanoid's proprioception, object states in the scene, and reference trajectories for object carrying.The action space directly outputs joint angles for the humanoid's entire body.

\textbf{Observation}:
Since we adopt a multi-stage RL training setup, each stage has distinct observation spaces and reward functions.
In the first stage, we focus on training the humanoid to rigorously follow the reference trajectory and transport objects to designated positions.
Stage 1 serves as the teacher model, where the observation space incorporates richer reference trajectory information.

In the teacher model training, we implement an asymmetric actor-critic architecture \cite{pinto2017asymmetric}.
The critic's observation space, beyond the standard observations, also includes the humanoid's body position and orientation from the reference motion.
Since we intend to test the teacher model in a motion capture room, the actor's observation space excludes the humanoid anchor's linear velocity, keeping it only in the critic.

\textbf{Rewards}:
In our task, the humanoid not only needs to track the specified motion trajectory, but also needs to move the object along the reference trajectory.
Here, we employ five reward functions, as detailed in Table \ref{tab:reward_setting}.

\begin{itemize}
\item \textit{Body Tracking}: DeepMimic-style tracking term for body position, orientation, linear and angular velocity.
\item \textit{Object Tracking}: tracking term for object position and orientation.
\item \textit{Guided Contact Reward}: tracking the relative position and orientation between the humanoid's palm and the object's center in the reference motion, and the contact state between the humanoid's hand and the object.
\item \textit{Regularization Penalties}: Self-Collision, Action Rate, Soft Joint Limit, joint acceleration, joint velocity limit, joint torque
\item \textit{Foot Sliding Penalty}: feet slide, feet contact consistency.
\end{itemize}

Both Body Tracking and Object Tracking reward functions adopt a Gaussian-shaped exponential error form, where smaller errors yield higher rewards.
\begin{equation}
    r(\bar{e}_s, \sigma_s) = \exp\left(-\bar{e}_s / \sigma_s^2\right)
\end{equation}

Unlike the BeyondMimic approach, our Body Tracking reward does not track all body segments.
Instead, we only track the position and orientation of key body parts such as the feet, legs, and upper arms.
By assigning different reward weights to key body parts, training converges faster and the humanoid learns the correct carrying behavior more easily.

\textbf{Terminations}:
Our termination conditions follow a similar design to BeyondMimic, 
where episodes are terminated when the humanoid's \textit{body tracking} error relative to the reference motion exceeds a predefined threshold.
We do not impose a termination condition for \textit{object tracking} errors 
because doing so would prevent the agent from ever encountering situations where the object deviates from the reference during training.
Without such exposure, the humanoid's motions could diverge dangerously 
when the object is lost or experiences significant deviation during real-world deployment. Detailed about Terminations can be found in Table~\ref{tab:Termination}.



\subsubsection{Stage 2: Assistant RL Training Setup for Enhanced Generalization}
A key objective of this work is to improve the generalization of humanoid object carrying tasks. 
Through few-shot learning to rapidly adapt to varied manipulation scenarios, 
we seek to develop more robust and generalizable carrying policies.

To enable the humanoid to transport objects at varying positions in the second stage, 
we modify the reward function by removing the global anchor position tracking reward 
and the humanoid's self-body position tracking reward.
We retain the body orientation tracking reward while increasing object position tracking and contact rewards.
This encourages the humanoid to learn to transport objects at different positions while maintaining proper body orientation.
The observations remain consistent with the first stage.
Regarding termiations, we increase the threshold for anchor position tracking error compared to stage 1.    
Note that removing the self-body position tracking reward may lead the humanoid to learn motions exceeding joint limits.
Therefore, we add a termination condition for joint angle limit violations.

This training setup reduces the humanoid's reliance on proprioceptive position and body position information, allowing it to focus more on object position and contact tracking.
This facilitates the distillation process in stage 3, where proprioceptive position observations are removed.

\subsubsection{Stage 3: Distillation of Proprioceptive Position Observations}
%
Intuitively, humans do not need to know their absolute position in the world coordinate system when carrying a object, only the object's position relative to the body is required.
Motivated by this observation, we perform distillation in the third stage to remove proprioceptive position observations from the policy input.
The resulting student policy retains only the minimal set of observations necessary for the task, which are summarized as follows:
\begin{itemize}
    \item Reference Motion: the anchor orientation error between the reference motion and the current humanoid state.
    \item Robot Proprioception: joint position, joint velocity, base angular velocity, and previous actions, with a history length of 5.
    \item Object Perception: object position, object orientation, and object contact point positions.
\end{itemize}
\textbf{We attempt to remove all reference motion guidance during the distillation stage, but this prevents the humanoid from determining the direction of the target, since after lifting the object, the object's relative pose no longer provides guidance to the humanoid.}

We adopt the DAgger-PPO algorithm for distillation, implemented following the rsl-rl codebase.
This algorithm combines a standard teacher-student distillation scheme with PPO.
At the beginning of training, the student policy primarily imitates the teacher's actions under the reduced observation set.
Once the imitation policy saturates the achievable reward, the PPO loss is introduced to further optimize the policy.
A KL divergence constraint is imposed to limit the distributional discrepancy between the student and teacher outputs, while the coefficients of the remaining PPO loss terms are kept consistent with stage 1.

The teacher provides PPO with a stable behavioral prior, and PPO further refines the teacher behavior using environment rewards, allowing the student to potentially exceed the teacher's performance.

\subsection{Anchor Pose Regression for Reference Motion-Free Deployment}

A policy conditioned on reference motion during training inevitably inherits a trajectory-following bias at deployment, preventing it from acquiring genuine generalization capability.

To completely eliminate the reference motion signal from the observation space at inference time, 
We design an LSTM-based \cite{vennerød2021longshorttermmemoryrnn} neural network, named \textit{QuatNet}, which is trained via supervised learning on the object's relative pose and the reference orientation of the humanoid's anchor point from the reference motion dataset.
\textbf{The inputs to the training set are the object's position and quaternion orientation relative to the humanoid, the humanoid's current joint positions, and the target transport position, while the label is the reference orientation of the humanoid's anchor point.}

\begin{figure}[htbp]
    \centerline{\includegraphics[width=9cm]{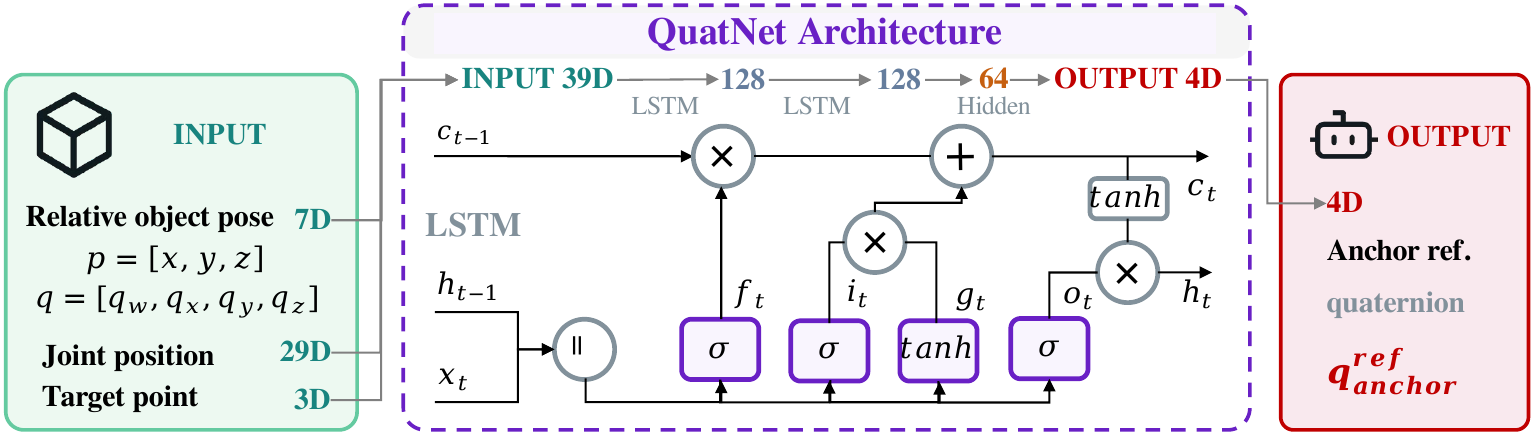}}
    \caption{\textbf{The architecture of the quaternion network.}
    The network takes as input the object's position and orientation relative to the humanoid, joint positions, target point,
    totaling 39 dimensions, and outputs the humanoid anchor reference orientation as a quaternion, 
    totaling 4 dimensions. It comprises two lstm layers and is trained 
    by minimizing the L2 distance between the predicted and ground-truth anchor reference quaternions.}
    \label{fig:3}
\end{figure}

The LSTM network infers the subsequent transport direction from the humanoid's current joint positions and the target position.The architecture of \textit{QuatNet} is illustrated in Fig~\ref{fig:3}.
Training is performed by minimizing the L2 distance between the predicted and ground-truth anchor orientation quaternions.
\begin{equation}
\mathcal{L}_{\text{quat}} = \left\| \mathbf{q}_{\text{pred}} - \mathbf{q}_{\text{gt}} \right\|_{2}^{2}
\end{equation}
At deployment, the humanoid perceives the object's relative pose via vision, and the learned network regresses the anchor reference pose to substitute for the reference motion term in the student policy's observation, 
yielding a policy that depends solely on real-time sensory input.

\subsection{Vision-based Object Pose Estimation}
At deployment, a fiducial tag attached to the object is detected by the humanoid's head-mounted RGB-D camera. Let $P$, $C$, $M$, and $O$ denote the pelvis, camera, tag, and object frames, respectively. The object pose in the pelvis frame is computed as
${}^{P}\mathbf{T}_{O}(t)
=
{}^{P}\mathbf{T}_{C}(\mathbf{q}_{t})
{}^{C}\mathbf{T}_{M}(t)
{}^{M}\mathbf{T}_{O},
$ 
where ${}^{C}\mathbf{T}_{M}(t)$ is obtained from tag detection, and ${}^{M}\mathbf{T}_{O}$ is fixed after the tag is mounted on the object. Since the head can rotate, ${}^{P}\mathbf{T}_{C}(\mathbf{q}_{t})$ is computed from the current head joint configuration $\mathbf{q}_{t}$ through forward kinematics, together with the calibrated camera extrinsics. The resulting object position and orientation in the pelvis frame are provided to the student policy.

%
To handle intermittent tag occlusion, we use a visibility-aware pose update that combines visual detection with hand kinematics. Each tag measurement is validated based on its reprojection error and consistency with recent observations before temporal smoothing. During a short loss of detection, the object pose is propagated using recent valid observations. 
Visual measurements are used again once the tag becomes visible.
\begin{table*}[ht]
    \caption{Comparison of success rate (Sr) and error (Err) across different baselines for different object positions and shape--size configurations.(pos* denotes the four positions of the object position offset, size1 denotes the cuboid, and size2 denotes the cylinder.)}
    \label{tab:1}
    \centering
    \begin{tabular}{@{}clcccccccccccccccc@{}}
    \toprule
    \multicolumn{2}{c}{\multirow{3}{*}{Method}} & \multicolumn{4}{c}{pos1}                              & \multicolumn{4}{c}{pos2}                              & \multicolumn{4}{c}{pos3}                              & \multicolumn{4}{c}{pos4}                              \\ \cmidrule(lr){3-6} \cmidrule(lr){7-10} \cmidrule(lr){11-14} \cmidrule(l){15-18}
    \multicolumn{2}{c}{}                        & \multicolumn{2}{c}{size1} & \multicolumn{2}{c}{size2} & \multicolumn{2}{c}{size1} & \multicolumn{2}{c}{size2} & \multicolumn{2}{c}{size1} & \multicolumn{2}{c}{size2} & \multicolumn{2}{c}{size1} & \multicolumn{2}{c}{size2} \\
    \multicolumn{2}{c}{}                        & Sr              & Err     & Sr              & Err     & Sr              & Err     & Sr              & Err     & Sr              & Err     & Sr              & Err     & Sr              & Err     & Sr             & Err      \\ \midrule
    \multicolumn{2}{c}{HDMI}                    & 85              & 0.079   & 88.9            & 0.076   & 83              & 0.105   & 92.6           & 0.074   & 75.4            & 0.097   & 84.2            & 0.088   & 58.2            & 0.107   & 76.4            & 0.092    \\
    \multicolumn{2}{c}{HumanX}                  & 0.0             & -       & 0.0             & -       & 0.0               & -       & 0.0              & -        & 0.0             & -       & 0.0             & -       & 0.0             & -       & 0.0             & -        \\
    \multicolumn{2}{c}{Ominiretarget}           & 79.7            & 0.147   & 80.5            & 0.310   & 11.3            & 0.505   & 3.9            & 1.951    & 24.4            & 0.751   & 14.5            & 3.080   & 2.9             & 1.323   & 9.2             & 3.628    \\
    \multicolumn{2}{c}{Stage 2}                 & \textbf{87.3}   & 0.075   & \textbf{92.0}   & 0.051   & \textbf{90.4}   & 0.069   & \textbf{90}    & 0.058   & \textbf{88.1}   & 0.090   & \textbf{89.6}   & 0.076   & \textbf{86.9}   & 0.084   & \textbf{88.7}   & 0.053    \\ 
    \multicolumn{2}{c}{Stage 3}                 & \textbf{94.3}   & 0.150   & 96.5   & 0.151   & \textbf{35.2}   & 0.334   & 55.7  & 0.286   & \textbf{58.4}   & 0.282   & 57.6   & 0.283   & \textbf{40.2}   & 0.327   & 32.0   & 0.356    \\
    \multicolumn{2}{c}{Ours}                    & \textbf{94.9}   & 0.151   & 95.9   & 0.151   & \textbf{21.9}   & 0.371   & 39.8  & 0.337   & \textbf{58.0}   & 0.290   & 62.1   & 0.273   & \textbf{31.2}   & 0.355   & 34.4   & 0.348    \\  \bottomrule
    \end{tabular}
\end{table*}

\begin{table*}[ht]
    \caption{Comparison of success rate (Sr) and error (Err) across training stages for each object position.}
    \label{tab:2}
    \centering
    \begin{tabular}{>{\centering\arraybackslash}m{0.05\linewidth}*{16}{>{\centering\arraybackslash}m{0.03\linewidth}}}        
    \toprule
    \multicolumn{2}{c}{\multirow{3}{*}{Method}} & \multicolumn{3}{c}{\multirow{2}{*}{w/o DA}}   & \multicolumn{3}{c}{\multirow{2}{*}{Stage 1}}     & \multicolumn{3}{c}{\multirow{2}{*}{\begin{tabular}[c]{@{}c@{}}Stage 2\\ (before distillation)\end{tabular}}} & \multicolumn{3}{c}{\multirow{2}{*}{\begin{tabular}[c]{@{}c@{}}Stage 3\\ (after distillation)\end{tabular}}} & \multicolumn{3}{c}{\multirow{2}{*}{\begin{tabular}[c]{@{}c@{}}Stage 3\\ (with QuatNet)\end{tabular}}} \\
    \multicolumn{2}{c}{}                        & \multicolumn{3}{c}{}                          & \multicolumn{3}{c}{}                               & \multicolumn{3}{c}{}                                                                                           & \multicolumn{3}{c}{}                                                                                    & \multicolumn{3}{c}{}                                                                                          \\ \cmidrule(l){3-17} 
    \multicolumn{2}{c}{}                        & Carry Sr & Grasp Sr & \multicolumn{1}{c}{Err} & Carry Sr      & Grasp Sr & \multicolumn{1}{c}{Err} & Carry Sr                          & Grasp Sr                     & \multicolumn{1}{c}{Err}                     & Carry Sr                     & Grasp Sr                    & \multicolumn{1}{c}{Err}                    & Carry Sr                         & Grasp Sr                       & \multicolumn{1}{c}{Err}                   \\ \midrule
    \multicolumn{2}{c}{pos1={[}0,0,0{]}}        & 90.4     & 91.8     & 0.067                   & \textbf{93.4} & 94.3     & 0.06                    & \textbf{93.4}                     & 93.7                         & 0.064                                       & \textbf{98.6}                & \textbf{100}                & 0.144                                     & \textbf{94.9}                    & \textbf{99.8}                   & 0.147                                     \\
    \multicolumn{2}{c}{pos2={[}0, -0.4, 0{]}}   & 87.9     & 93.6     & 0.122                   & 90.6          & 92.4     & 0.06                    & 91                                & 92                           & 0.066                                       & 44.1                         & 99.8                        & 0.321                                     & 29.1                             & 100                           & 0.360                                     \\
    \multicolumn{2}{c}{pos3={[}-0.25,0.25,0{]}} & 82.4     & 91.6     & 0.201                   & \textbf{84.4} & 90.2     & 0.12                    & \textbf{87.7}                     & 90.2                         & 0.085                                       & \textbf{59.6}                         & \textbf{99.2}                        & 0.286                                     & 57.4                             & 98.4                           & 0.281                                     \\ 
    \multicolumn{2}{c}{pos4={[}0.25,0.45,0{]}}  & 73.4     & 89.3     & 0.351                   & 68.6 & 88.1     & 0.345                   & 89.5                     & 91.2                         & 0.071                                       & 53.5                         & 99.8                        & 0.296                                     & 32.4                             & 99.0                           & 0.351                                     \\ \bottomrule
    \end{tabular}
\end{table*}

\section{Experiments}

\subsection{Experimental Setup}
We collected 100 video sequences and applied data augmentation to generate 4000 motion trajectories with varied initial object positions 
and object shape--size configurations. The augmented motion trajectories were retargeted to the 29Dof Tiangong 3.0 humanoid for policy training. 
\textit{Isaac Lab} was used as the training simulator, while \textit{MuJoCo} was used for all simulation-based evaluations.

We first compare our method with existing approaches, including HDMI, HumanX, and OmniRetarget, on a object-carrying task. 
We report both the task success rate and the corresponding error under four object placements and two object shape--size configurations. 
A trial is considered successful when the humanoid transports the object to within 0.3 m of the target position.
We design ablation experiments in simulation to evaluate the impact of all components on performance.

\subsection{Comparison with other baseline methods}
As shown in Table \ref{tab:1}, because the tracking of body position in the reference motion is weakened, our two-stage method achieves a significantly higher carrying success rate than other methods under the condition of object position generalization, and it can adapt to objects of different shapes and sizes.To make our policy more generalizable and deployable, we incorporated the distillation method and the \textit{QuatNet}. With the introduction of these two mechanisms, however, the success rate of the policy decreases due to the lack of guidance from the reference motion.

\subsection{Results of ablation experiments}
Following the MOVER training pipeline, we compare five policies trained under different configurations: 
(1) without data augmentation, (2) stage-1 teacher policy, (3) stage-2 assistant policy, (4) stage-3 student policy, (5) student policy with QuatNet.
We evaluate all policies across four different object positions in simulation, measuring three metrics: grasping success rate (whether the object is successfully lifted by the humanoid), carrying success rate (whether the object is delivered to the vicinity of the target location), 
and carrying error (the Euclidean distance between the final object position and the target).

As shown in Table \ref{tab:2}, across varying object position offsets, the pick-and-place success rate decreases progressively, with larger offsets leading to more pronounced degradation. Meanwhile, training strategies with data augmentation consistently achieve higher success rates compared to those without augmentation.

When there is no position offset, the stage-1 teacher policy and stage-2 assistant policy achieve similar carrying success rates. However, as the offset grows, the assistant policy significantly surpasses the teacher policy, demonstrating its superior position generalization capability. 
When the student policy is distilled to discard proprioceptive position information, its carrying success rate under large position offsets drops noticeably below that of the assistant policy, suggesting that proprioceptive position cues assist the humanoid in estimating the carrying target location. 
Moreover, after DAgger-PPO distillation, the student policy consistently achieves higher grasping success rates than both the teacher and assistant policies. 
This is mainly because the student distills away part of the reference information, allowing the humanoid to obtain strong guidance from the object position information.

\begin{figure}[htbp]
    \centerline{\includegraphics{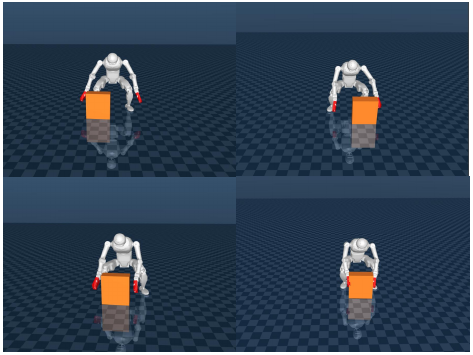}}
    \caption{Comparison of the grasping effects of Stage 1 and Stage 2.
    We set two object placement positions and compare the grasping effects of Stage 1 and Stage 2. 
    (Top) shows the Stage-1 policy, and (Bottom) shows the Stage-2 policy.
    }
    \label{fig:4}
\end{figure}

We further conducted a comparative analysis of the grasping effects between the stage-1 and stage-2 policies in MuJoCo,
as shown in Fig~\ref{fig:4}. Under conditions of significant object position offset, 
the stage-1 policy struggles to accurately guide the humanoid to the front of the object's placement location, 
owing to its dependence on the reference motion trajectory. 
In contrast, the stage-2 policy exhibits a notable ability to track the object's position throughout the manipulation.

\subsection{QuatNet experiment result}
\begin{figure}[htbp]
    \centerline{\includegraphics{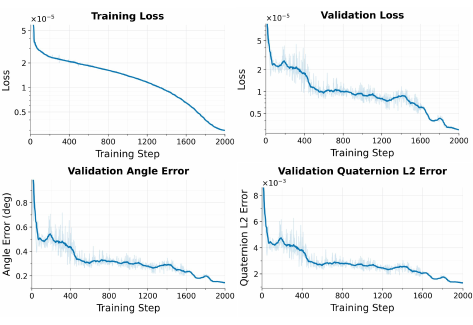}}
    \caption{QuatNet training curves. The experiment records the training set loss and validation set loss, 
    as well as the validation set loss represented by the rotation angle and quaternion difference.}
    \label{fig:5}
\end{figure}

During the training of the \textit{QuatNet} network, we use 500 motion sequences, yielding a dataset of approximately 1,000,000 frames, 
with 20\% reserved for validation. The model is trained for 2000 epochs, during which we record the training loss, validation loss, as well as the rotation angle error and quaternion difference on the validation set. 
After training, the output error meets the required precision threshold, as shown in Fig~\ref{fig:5}.

From the last column of Table \ref{tab:2}, the policy that uses \textit{QuatNet} to replace the reference motion observation achieves a lower carrying success rate than the student policy, but a higher success rate than the policy that uses neither the reference motion nor \textit{QuatNet}, which success rate is typically below 10\%.

\subsection{Real-world test result}
For real-world evaluation, we use a motion-capture room to measure the global poses of the humanoid and the object. 
These measurements are provided as observations to evaluate the teacher policy. 
In a conventional indoor environment, an RGB-D camera estimates the object's pose relative to the humanoid, 
and these relative observations are used to evaluate the student policy. 
We place the object at three different locations and compare the carrying success rates between the motion-capture-room and conventional-room settings.

As can be seen from Table \ref{tab:3}, the success rate of the tests conducted in the motion capture room is higher than that of the tests conducted under visual conditions, and the carrying error is also smaller, while both demonstrate strong generalization ability to objects at different positions and can maintain stable performance in multiple consecutive tests.

\begin{table}[t]
    \caption{Comparison of success rate (Sr) and error (Err) before and after distillation in real-world tests with Mocap and Vision observations.}
    \label{tab:3}
    \centering
    \begin{tabular}{@{}lllll@{}}
    \toprule
    \multicolumn{1}{c}{\multirow{3}{*}{Method}} & \multicolumn{2}{c}{\multirow{2}{*}{\begin{tabular}[c]{@{}c@{}}Stage 2\\ (with Mocap)\end{tabular}}} & \multicolumn{2}{c}{\multirow{2}{*}{\begin{tabular}[c]{@{}c@{}}Stage 3\\ (with Vision)\end{tabular}}} \\
    \multicolumn{1}{c}{}                        & \multicolumn{2}{c}{}                                                                                  & \multicolumn{2}{c}{}                                                                                   \\ \cmidrule(lr){2-3} \cmidrule(lr){4-5} 
    \multicolumn{1}{c}{}                        & Sr                                                & Err                                               & Sr                                                & Err                                                \\ \midrule
    pos1                                        & 7/10                                              & 0.18                                              & 7/10                                              & 0.20                                                    \\
    pos2                                        & 6/10                                              & 0.27                                              & 5/10                                              & 0.36                                                   \\
    pos3                                        & 6/10                                              & 0.34                                              & 5/10                                              & 0.41                                                   \\ \bottomrule
    \end{tabular}
\end{table}

\section{Conclusion}


In this work, we presented \textbf{MOVER}, a multi-stage whole-body interaction training framework for humanoid object-carrying.
We first retargeted video data into humanoid reference trajectories and injected augmented object trajectories into the retargeted data.
A teacher-assistant-student whole-body interaction imitation learning pipeline was then performed on these trajectories.
Through the assistant policy training, the humanoid learns to transport objects placed at varying positions 
and can complete carrying tasks even when global position information from a motion-capture system is unavailable.
Furthermore, we designed a \textit{QuatNet} network that, conditioned on the current object's relative pose, 
replaces the reference motion term in the policy observation.
This enabled reference-motion-free object interaction in simulation.
In simulated object-carrying experiments with varying object positions, our approach achieved a higher success rate than existing methods.
Finally, we deployed the policy on a physical humanoid and successfully completed the prescribed task in a motion-capture-free setting.



\bibliographystyle{IEEEtran}
\bibliography{ref}

\section{Appendix}
\subsection{PPO Hyperparameters}
The hyperparameters employed in Stages 1 and 2 are summarized in Table~\ref{tab:ppo_hyperparameters}, 
while the DAggerPPO hyperparameters used in Stage~3 are reported in Table~\ref{tab:dagger_ppo_hyperparameters}.
During DAggerPPO distillation, the policy is first trained with DAgger-based imitation learning for the first 3000 steps (controlled by \textit{ppo start step}); 
thereafter, the PPO loss is introduced to continue training.

\subsection{Observation space}
The observations and their corresponding dimensions at the different stages—teacher, assistant, and student—can be found in Table~\ref{tab:teacher_assistant_student_obs}.

\subsection{Reward Setting}
\label{sec:reward_setting}
The complete reward configuration used during training is summarized in
Table~\ref{tab:reward_setting}. Rewards are organized into three groups:
\textit{Tracking} rewards that drive the robot to imitate the reference motion,
\textit{Interaction} rewards that encourage correct manipulation of the object,
and \textit{Regularization} rewards that regularize the policy for smoothness and safety.

\begin{table}[htbp]
\caption{PPO hyperparameters used in our experiments.}
\label{tab:ppo_hyperparameters}
\centering
\begin{tabular}{lc}
\toprule
\textbf{Hyperparameter} & \textbf{Value} \\
\midrule
num steps per env     & 24 \\
max iterations         & 100000 \\
init noise         & 1.0 \\
actor hidden dims      & [512, 256, 128] \\
critic hidden dims     & [512, 256, 128] \\
activation              & elu \\
\midrule
value loss coef        & 1.0 \\
use clipped value loss & True \\
clip param             & 0.2 \\
entropy coef           & 0.005 \\
num learning epochs    & 5 \\
num mini batches       & 4 \\
learning rate          & 1.0$\times$10$^{-3}$ \\
\midrule
gamma                   & 0.99 \\
lamda                     & 0.95 \\
desired kl             & 0.01 \\
max grad norm         & 1.0 \\
\bottomrule
\end{tabular}
\end{table}

\begin{table}[htbp]
\caption{DAggerPPO hyperparameters used in our experiments.}
\label{tab:dagger_ppo_hyperparameters}
\centering
\begin{tabular}{lc}
\toprule
\textbf{Hyperparameter} & \textbf{Value} \\
\midrule
num steps per env     & 24 \\
max iterations         & 50000 \\
init noise std        & 0.05 \\
actor hidden dims      & [512, 256, 128] \\
critic hidden dims     & [512, 256, 128] \\
teacher hidden dims    & [512, 256, 128] \\
activation              & elu \\
\midrule
value loss coef        & 1.0 \\
entropy coef           & 0.005 \\
ppo loss coef          & 0.1 \\
ppo start step         & 3000 \\
dagger loss coef       & 1.0 \\
dagger loss type       & mse \\
learning rate          & 1.0$\times$10$^{-3}$ \\
max grad norm         & 1.0 \\
desired kl             & 0.01 \\
\bottomrule
\end{tabular}
\end{table}

\begin{table}[htbp]
    \centering
    \caption{Domain Randomization Parameters}
    \label{tab:domain-randomization}
    \begin{tabular}{ll}
    \toprule
    \textbf{Domain Randomization} & \textbf{Range} \\
    \midrule
    Object friction coefficient & (0.4, 0.8) \\
    Object mass (kg) & (0.8, 3.5) \\
    Object initial position (m) & (-0.15, 0.15) \\
    Initial RPY angle (rad) & (-0.2, 0.2) \\
    Object initial linear velocity (m/s) & (-0.06, 0.06) \\
    Object initial angular velocity (rad/s) & (-0.05, 0.05) \\
    Joint friction coefficient & (0.001, 0.6) \\
    Armature coefficient & (0.002, 0.06) \\
    Rigid body friction coefficient & (0.3, 1.6) \\
    Default joint angle offset (rad) & (-0.01, 0.01) \\
    Base center of mass: x (m) & (-0.025, 0.025) \\
    Base center of mass: y (m) & (-0.05, 0.05) \\
    Base center of mass: z (m) & (-0.05, 0.05) \\
    Body mass factor & (0.7, 1.3) \\
    PD controller parameter & (0.75, 1.25) \\
    Random push velocity (m/s) & (1.0, 5.0) \\
    \bottomrule
    \end{tabular}
\end{table}

\begin{table}[htbp]
  \centering
  \caption{Termination conditions and thresholds}
  \label{tab:Termination}
  \begin{tabular}{ll}
    \toprule
    \textbf{Termination} & \textbf{threshold} \\
    \midrule
    motion\_done & -- \\
    joint\_limit & -- \\
    anchor\_pos\_z & 0.15 \\
    anchor\_pos & 0.5 \\
    anchor\_ori & 0.8 \\
    ee\_body\_pos & 0.25 \\
    \bottomrule
  \end{tabular}
\end{table}

\begin{table}[htbp]
    \caption{Observation spaces used by the Teacher/Assistant policy and the Student policy.}
    \label{tab:teacher_assistant_student_obs}
    \centering
    \begin{tabular}{lc lc}
    \toprule
    \multicolumn{2}{c}{\textbf{Teacher / Assistant}} & \multicolumn{2}{c}{\textbf{Student}} \\
    \midrule
    \textbf{Item} & \textbf{Dim} & \textbf{Item} & \textbf{Dim} \\
    \midrule
    command joint pos       & $29$       & motion anchor ori b   & $6$ \\
    command joint vel       & $29$       & base ang vel          & $3 \times 5$ \\
    motion anchor pos b     & $3$        & joint pos             & $29 \times 5$ \\
    motion anchor ori b     & $6$        & joint vel             & $29 \times 5$ \\
    base ang vel            & $3$        & actions               & $29 \times 5$ \\
    joint pos               & $29$       & object position       & $3$ \\
    joint vel               & $29$       & object orientation    & $4$ \\
    actions                 & $29$       & object contact pos    & $6$ \\
    object position         & $3$        &                       & \\
    object orientation      & $4$        &                       & \\
    target obj pos          & $3$        &                       & \\
    target obj quat         & $4$        &                       & \\
    ref contact pos b       & $6$        &                       & \\
    object contact pos      & $6$        &                       & \\
    \midrule
    \multicolumn{2}{l}{\emph{Privileged observations}} & & \\
    \midrule
    body pos                & $96$       &                       & \\
    body ori                & $192$      &                       & \\
    base lin vel            & $3$        &                       & \\
    \bottomrule
    \end{tabular}
\end{table}

\begin{table}[htbp]
    \caption{Reward functions and their corresponding weights. Positive weights promote the desired behavior, while negative weights penalize undesired effects.}
    \label{tab:reward_setting}
    \centering
    \resizebox{\linewidth}{!}{%
    \begin{tabular}{lc@{\hspace{2ex}}lc}
    \toprule
    \textbf{Reward} & \textbf{Weight} & \textbf{Reward} & \textbf{Weight} \\
    \midrule
    \multicolumn{2}{l}{\emph{Robot Tracking}} & 
    \multicolumn{2}{l}{\emph{Object Tracking}} \\
    \midrule
    motion global anchor pos       & $1.0$         & object position                 & $2.0$ \\
    motion global anchor ori      & $0.5$         & object orientation              & $2.0$ \\
    foot body pos                  & $2.0$         & body object relative pos         & $3.0$ \\
    foot body ori                  & $1.5$         & body object relative ori         & $1.0$ \\
    leg body pos                   & $1.0$         & contact graph                   & $2.0$ \\
             
    \cmidrule{3-4}
    \multicolumn{2}{l}{} & \multicolumn{2}{l}{\emph{Regularization Penalties}} \\
    \cmidrule{3-4}
    leg body ori                   & $0.5$         & action rate l2                  & $-5.0\times10^{-1}$ \\
    upper arm body pos             & $1.0$         & joint torque l2                 & $-1.0\times10^{-5}$ \\
    upper arm body ori             & $0.8$         & joint acc                       & $-2.5\times10^{-7}$ \\
    elbow body pos                 & $1.5$         & joint limit                     & $-30.0$ \\
    elbow body ori                 & $2.0$         & joint vel limit                 & $-0.1$ \\
    wrist body pos                 & $1.5$         & undesired contacts              & $-0.1$ \\
    wrist body ori                 & $2.0$         & feet slide                      & $-2.5$ \\
    finger body pos                & $1.5$         &                                 & \\
    finger body ori                & $1.0$         &                                 & \\
    motion body lin vel            & $1.0$         &                                 & \\
    motion body ang vel            & $1.0$         &                                 & \\
    foot body lin vel              & $1.5$         &                                 & \\
    foot contact consistency        & $1.5$         &                                 & \\
    \bottomrule
    \end{tabular}%
    }
\end{table}

\end{document}